\documentclass{article}
\usepackage{arxiv}
\usepackage{graphicx, float}%
\usepackage{multirow}%
\usepackage{amsmath,amssymb,amsfonts}%
\usepackage{amsthm}%
\usepackage{mathrsfs}%
\usepackage[title]{appendix}%
\usepackage{xcolor}%
\usepackage{textcomp}%
\usepackage{manyfoot}%
\usepackage{booktabs}%
\usepackage{algorithm}%
\usepackage{algorithmicx}%
\usepackage{algpseudocode}%
\usepackage{listings}%
\usepackage{array}
\usepackage[position=top]{subcaption}
\usepackage{dcolumn}
\usepackage[utf8]{inputenc} 
\usepackage{soulutf8}
\usepackage{soul}
\usepackage{wrapfig}
\usepackage{soul, color}

\begin{document}
%

\title{Near-Infrared and Low-Rank Adaptation of\\Vision Transformers in Remote Sensing}
%
%
%

\author{Irem~Ulku$^1$,
        O. Ozgur~Tanriover$^1$,
        ~Erdem~Akagündüz$^2$, \\
         $^1$Dept. of Computer Science, Ankara University, Ankara,
Türkiye\\
    $^2$Dept. of Modeling and Simulation, Graduate School of Informatics, METU, Ankara, Türkiye \\
	\texttt{\{irem.ulku,tanriover\}@ankara.edu.tr, akaerdem@metu.edu.tr}} 
%
%

\markboth{Journal of Remote Sensing Letters,~Vol.~?, No.~?, ???~2024}%
{Shell \MakeLowercase{\textit{et al.}}: Bare Demo of IEEEtran.cls for Journals}
%



\maketitle

\begin{abstract}
Plant health can be monitored dynamically using multispectral sensors that measure Near-Infrared reflectance (NIR). Despite this potential, obtaining and annotating high-resolution NIR images poses a significant challenge for training deep neural networks. 
Typically, large networks pre-trained on the RGB domain are utilized to fine-tune infrared images. This practice introduces a domain shift issue because of the differing visual traits between RGB and NIR images. 
As an alternative to fine-tuning, a method called low-rank adaptation (LoRA) enables more efficient training by optimizing rank-decomposition matrices while keeping the original network weights frozen. However, existing parameter-efficient adaptation strategies for remote sensing images focus on RGB images and overlook domain shift issues in the NIR domain. Therefore, this study investigates the potential benefits of using vision transformer (ViT) backbones pre-trained in the RGB domain, with low-rank adaptation for downstream tasks in the NIR domain. Extensive experiments demonstrate that employing LoRA with pre-trained ViT backbones yields the best performance for downstream tasks applied to NIR images.
\end{abstract}


%

\section{Introduction}
%
%
%
%
The near-infrared (NIR) spectrum offer the most valuable insights into the health of vegetation on Earth \cite{gkillas2023cost}. For example, when a plant is attacked by a pest or under stress, there's a noticeable change in the reflectance of NIR bands due to alterations in the internal leaf structure. NIR reflectance can uniquely identify the internal leaf structure and biochemical composition, while also distinguishing healthy plants. Therefore, multispectral sensors covering NIR bands have the capability to automatically monitor vegetation health \cite{arogoundade2023leveraging}. However, NIR sensors typically possess lower spatial resolutions and lower signal-to-noise ratios compared to RGB sensors \cite{aslahishahri2021rgb}. Consequently, acquiring and annotating a sufficient number of NIR images with high resolution and signal-to-noise ratio to effectively train deep neural networks is a state-of-the-art challenge \cite{zhang2023two}.

A common practice to address this problem is to train deep neural networks with RGB images and transferring these pre-trained networks with infrared images \cite{ma2024transfer}. However, the visual characteristics of RGB images exhibit distinct distributions compared to infrared images, resulting in the widely recognized "domain shift problem". This issue in deep learning is often addressed by fine-tuning the initial model using inputs from the shifted domain. Depending on factors such as model complexity, domain shift scale, and dataset size, solving this problem can prove achievable to a certain extent. The reader may refer to \cite{aslahishahri2021rgb} for a review of solutions to the domain shift problem in deep learning. 

Currently, fine-tuning is realized mainly by foundation models, which require large-scale data for training. Among various foundation models, transformer-based architectures \cite{dosovitskiy2020image, liu2021swin} offer the best solution, since they achieve lower inductive bias than convolutional neural networks with the help of positional embeddings \cite{ma2024transfer}. Although transformer-based models can improve performance, the benefit comes with the cost of the highest computational burden. This drawback primarily stems from the requirement of large-scale pretraining on supervised datasets. As a still ongoing field, transformer-based foundation models have a high potential for usage in remote sensing-based tasks to support domain adaptation. To effectively utilize large foundation models for multispectral remote-sensing-based agriculture tasks, the models must address difficult challenges, such as achieving high accuracy in real-time speed. The difficulty mainly comes from the need for a higher number of trainable parameters to meet the accuracy requirements, consequently slowing down both training and inference speeds. These constraints render foundation models impractical for integration into real-time mobile applications. \cite{olimov2021ref}. 

There is a recent trend towards  adaptation methods instead of fine-tuning, as they offer improved generalization in out-of-domain scenarios due to avoiding catastrophic forgetting, especially in cases of limited data availability \cite{li2023enhancing}. 
These family of methods are referred to as ``adaptation methods'',  which tune only few parameters to mitigate overfitting, and can alleviate the storage burden and increase the performance of traditional fine-tuning approaches \cite{he2023sensitivity}. A recent example of this family, namely Low-Rank Adaptation (LoRA) \cite{hu2021lora} approach allows for indirect training of dense layers while freezing original weights by only optimizing the rank-decomposition matrices of the changes in these layers. The underlying hypothesis is that change in weight matrices during adaptation has a low intrinsic rank. Although simple and computationally efficient, its usage remains confined to the transformer-based large language models (LLM) from which it originated, with limited research on vision transformers \cite{liu2022polyhistor}.


Few studies exist in recent literature that analyze the parameter-efficient adaptation strategies with remote-sensing images \cite{xue2024adapting, hu2024airs, dong2024upetu}. These studies solely focus on adapting RGB images and do not delve into the challenges associated with domain shift when adapting to the NIR domain. To address this gap, this paper proposes a framework to investigate the impact of low-rank adaptation on the semantic segmentation performance of vision transformers (ViT) on NIR images. 
To the best of our knowledge, near-infrared and low-rank adaptation of vision transformers has not been used for the semantic segmentation of multispectral images in remote sensing.


\begin{figure*}[t]
    \centering
    \includegraphics[trim=0 0 0 0,clip=true,width = 0.99\textwidth]{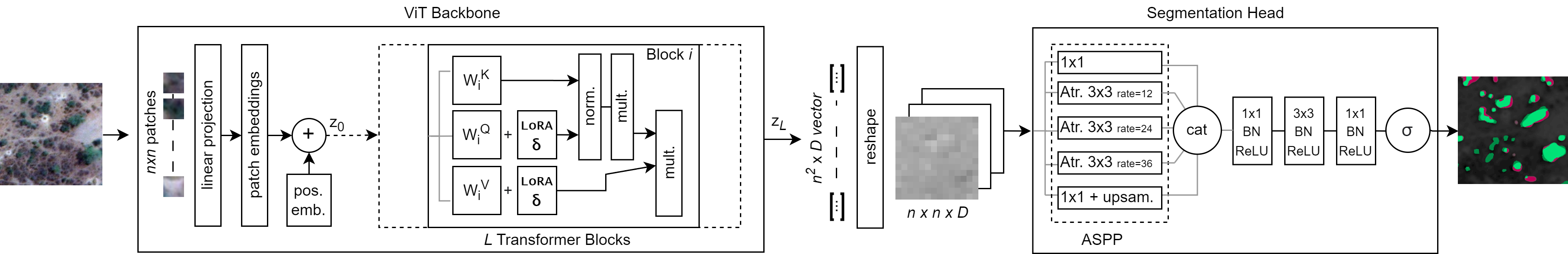}
    \caption{The proposed LoRA-based ViT+segmentation acrhictecture. }
    \label{fig:arch}
\end{figure*}

\section{Model Architecture}

\subsection{Encoder}
Fig.\ref{fig:arch} shows an embedding layer and an encoder component based on the ViT architecture. The input image $\mathbf{x}\in \mathbb{R}^{H\times W\times C}$ with dimensions as height $H$, width $W$ and channel $C$, is divided into a sequence of $\mathbf{x}=\left [ x_{1}, ...,x_{N} \right ]\in\mathbb{R}^{N\times P^{2}\times C}$  patches, each with dimensions $\left ( P,P \right ) $, resulting in a total of $N=HW/P^{2}$ patches. Each patch gets flattened into a one-dimensional vector, and these flattened patches are mapped onto a sequence of patch embeddings $\mathbf{x}_{0}=\left [ \mathbf{E}x_{1},...,\mathbf{E}x_{N} \right ]\in \mathbb{R}^{N\times D}$ in a D-dimensional latent space using a linear projection layer $\mathbf{E}\in \mathbb{R}^{D\times \left ( P^{2}C \right )} $. To capture spatial relationships, learnable positional embeddings $\mathbf{E}_{pos}\in \mathbb{R}^{ N\times D}$ are added to the projected patch embedding sequence as follows:
\begin{equation}
\label{eq1}
 \mathbf{z}_{0}=\left [ \mathbf{E}x_{1},...,\mathbf{E}x_{N} \right ]+\mathbf{E}_{pos}.
\end{equation}

The contextualized encodings sequence $\mathbf{z}_{L}\in \mathbb{R}^{N\times D}$, obtained by applying $L$ transformer blocks to the token array $\mathbf{z}_{0}$, is formed. Several transformer blocks are stacked to obtain latent embeddings $z_{i}$, where each transformer block consists of a multi-headed self-attention (MHA) block followed by a multilayer perceptrons (MLP) block:

\begin{equation}
\label{eq2}
{z}'_{i}=MHA\left ( Norm\left ( z_{i-1} \right ) \right )+ z_{i-1} 
\end{equation}

\begin{equation}
\label{eq3}
{z}_{i}=MLP\left ( Norm\left ( {z}'_{i-1} \right ) \right )+ {z}'_{i-1},
\end{equation}
where $i=1,...,L$ and $Norm\left ( . \right )$ represents layer normalization.

The self-attention mechanism involves the query $\mathbf{Q}_{i}\in \mathbb{R}^{N\times d}$, key $\mathbf{K}_{i}\in \mathbb{R}^{N\times d}$, and value $\mathbf{V}_{i}\in \mathbb{R}^{N\times d}$ by multiplying the patch embedding matrix $\mathbf{z}_{0}\in \mathbb{R}^{N\times D}$ with weight matrices:
\begin{equation}
\label{eq4}
\mathbf{Q}_{i}=\mathbf{z}_{0}\mathbf{W}^{Q}_{i},\mathbf{K}_{i}=\mathbf{z}_{0}\mathbf{W}^{K}_{i},\mathbf{V}_{i}=\mathbf{z}_{0}\mathbf{W}^{V}_{i},
\end{equation}
where $\mathbf{W}^{Q}_{i},\mathbf{W}^{K}_{i},\mathbf{W}^{V}_{i}$ are learnable weight matrices. When the MHA block contains $L$ attention heads, the self-attention operation for the $i^{th}$ head is as follows:
\begin{equation}
\label{eq5}
Atten_{i}\left ( Q,K,V \right )=softmax\left ( \mathbf{Q}_{i}\mathbf{K}_{i}^{T} /\sqrt{D}\right )\mathbf{V}_{i},
\end{equation}

The MHA mechanism which identifies the regions to give attention using cosine similarity then projects these concatenated heads through another learnable parameter, denoted as $\mathbf{W}\in \mathbb{R}^{D\times D}$:
\begin{equation}
\label{eq7}
MHA\left ( \mathbf{z}_{L} \right )=\left [ head_{1},...,head_{L} \right ]\mathbf{W}.
\end{equation}

These series of transformer mappings align the sequence activation $\mathbf{z}_{i-1}=\left [ z_{i-1,1},...,z_{i-1,N} \right ]$, to a contextualized encoded sequence $\mathbf{z}_{i}=\left [ z_{i,1},...,z_{i,N} \right ] $ with enriched semantic information.

\subsection{LoRA Adaptation}
The LoRA layers are applied to the query and value projection layers in each transformer block based on the hypothesis that the updates of weights have a low intrinsic rank during adaptation. Assuming the pre-trained weight matrix of any projection layer as $\mathbf{W}\in \mathbb{R}^{C_{in}\times C_{out}}$, the updated version incorporates two linear layers, denoted as $\mathbf{A}\in \mathbb{R}^{r\times C_{out}}$ and $\mathbf{B}\in \mathbb{R}^{C_{in}\times r}$, while $W$ remains frozen to avoid receiving gradient updates according to:
\begin{equation}
\label{eq8}
h=\mathbf{W}x+\bigtriangleup \mathbf{W}x=\mathbf{W}x+\mathbf{BA}x.
\end{equation}
The value of $r$, such that $r\leqslant min\left ( C_{in},C_{out} \right )$, indicates a substantial reduction in the trainable parameters. The resulting output vectors are summed element-wise after multiplying $\mathbf{W}$ and $\bigtriangleup \mathbf{W}$ with the same input. It is reasonable to apply the LoRA to the query and value projection layers to adjust the attention scores within the MHA block:
\begin{equation}
\label{eq9}
\mathbf{Q}_{i}=\mathbf{z}_{0}\mathbf{W}^{Q}_{i}+\mathbf{z}_{0}\mathbf{B}^{Q}_{i}\mathbf{A}^{Q}_{i}
\end{equation}
\begin{equation}
\label{eq10}
\mathbf{V}_{i}=\mathbf{z}_{0}\mathbf{W}^{V}_{i}+\mathbf{z}_{0}\mathbf{B}^{V}_{i}\mathbf{A}^{V}_{i}
\end{equation}
\begin{equation}
\label{eq11}
\mathbf{K}_{i}=\mathbf{z}_{0}\mathbf{W}^{K}_{i}
\end{equation}
where $\mathbf{W}^{Q}_{i}$, $\mathbf{W}^{V}_{i}$ and $\mathbf{W}^{K}_{i}$ represent frozen weights obtained from ViT, while $\mathbf{B}^{Q}_{i}$, $\mathbf{A}^{Q}_{i}$, $\mathbf{B}^{V}$ and $\mathbf{A}^{V}_{i}$ denote trainable LoRA parameters.
 
\subsection{Segmentation Head}
ASPP (Atrous Spatial Pyramid Pooling) \cite{chen2017deeplab} is used to design a lightweight segmentation head that captures multiple scales of context information. ASPP consists of five parallel branches so that while providing feature mapping with $1\times 1$ convolution, three atrous convolutions with $\left [ 12, 24, 36 \right ]$ atrous rates enable multi-scale feature extraction. ASPP pooling with an adaptive average pooling layer and $1\times 1$ convolution can capture global information, while bilinear interpolation recovers feature size. As a result of concatenating all these feature maps into the channel dimension and fusing them with $1\times 1$ convolution, the output feature map is passed through the Sigmoid function for binary segmentation after $3\times 3$ and $1\times 1$ convolutions.

\section{Experimental Results}

\subsection{Image sets}
The DSTL image set \cite{ulku2022deep} comprises 25 satellite images covering a region of 1000 m × 1000 m. RGB images have dimensions of 3348 × 3392 pixels with a spatial resolution of 0.31 m. Multispectral images, on the other hand, have dimensions of 837 × 848 pixels and a spatial resolution of 1.24 m, spanning the wavelength range of 400-1040 nm. The DSTL image set focuses on the crop target class.

The RIT-18 image set \cite{ulku2022deep} contains a single aerial training image with a spatial resolution of 0.047 m and dimensions of 9393 × 5642 pixels. RIT-18 provides multispectral images at wavelengths of 715-725, 795-805, and 890-910 nm and assesses the tree target class.

Images from both sets are cropped into 224 × 224 patches and split into $72\%$ for training, 20\% for testing, and 8\% for validation.

\begin{table}[htbp]
\caption{Backbone details}
\begin{center}
\begin{tabular}{llcccccl}
 &                   & \multicolumn{1}{l}{} & \multicolumn{1}{l}{} & \multicolumn{1}{l}{} & \multicolumn{1}{l}{} & \multicolumn{1}{l}{} &  \\ \cline{2-7}
 & \textbf{Backbone} & \textbf{Patch} & \textbf{Layers} & \textbf{Hidden} & \textbf{MLP} & \textbf{Heads} &  \\ \cline{2-7}
 & B\_16 & 16                   & 12                   & 768                  & 3072                 & 12                   &  \\
 & B\_32 & 32                   & 12                   & 768                  & 3072                 & 12                   &  \\
 & L\_16 & 16                   & 24                   & 1024                 & 4096                 & 16                   &  \\
 & L\_32 & 32                   & 24                   & 1024                 & 4096                 & 16                   &  \\ \cline{2-7}
 &                   & \multicolumn{1}{l}{} & \multicolumn{1}{l}{} & \multicolumn{1}{l}{} & \multicolumn{1}{l}{} & \multicolumn{1}{l}{} & 
\end{tabular}%
\label{Tab:tab1}
\end{center}
\end{table}

\subsection{Experimental Setup}
The experiments are implemented on NVIDIA Quadro RTX 5000 GPU using the Pytorch framework. Training employs the Adam algorithm for 70 epochs with a mini-batch size of 8 and an initial learning rate of 1e-4. Learning rate is reduced by 9\% in every five iterations. The loss function for all experiments is binary cross-entropy with logits.

The chosen metrics for performance evaluation are the Jaccard index (IoU) and the $F_{1}$ score. The Jaccard index calculation is  $IoU = (TP / (TP + FP + FN))$, where $TP$, $FP$, and $FN$ are the number of true positive, false positive, and false negative pixels, respectively. The $F_{1}$ score is the harmonic mean of precision (p) and recall (r) metrics. The calculation of precision is $p = (TP / (TP + FP)) $, and that of the recall is $r = (TP / (TP + FN)) $.

\subsection{Results and Discussions}

\begin{table*}[h]
\centering
\caption{Tree semantic segmentation test results in terms of Jaccard Index (IoU) and F\textsubscript{1} score for the different architectures with RIT-18 image set}\label{tab1}
\resizebox{\textwidth}{!}{%
\begin{tabular}{cc|c|cccc}
\multicolumn{1}{l}{\textbf{}} &
  \multicolumn{1}{l|}{} &
  \multirow{2}{*}{\textbf{Trainable Parameters(M)}} &
  \multicolumn{2}{c}{\textbf{RGB}} &
  \multicolumn{2}{c}{\textbf{NIR}} \\ \cline{4-7} 
\textbf{Backbone}                                       & \textbf{Model}             &         & \textbf{IoU}  & \textbf{F1}   & \textbf{IoU}  & \textbf{F1}   \\ \hline
\multicolumn{1}{c|}{Xception}                           & DeepLabV3                  & 58.80   & 0.837 ± 0.318 & 0.859 ± 0.303 & 0.857 ± 0.298 & 0.878 ± 0.282 \\ \hline
\multicolumn{1}{c|}{\multirow{2}{*}{B\_16\_imagenet1k}} & ViT + DeepLabV3 Head       & 91.49  & 0.859 ± 0.283 & 0.885 ± 0.260 & 0.881 ± 0.258 & 0.905 ± 0.236 \\
\multicolumn{1}{c|}{}                                   & LoraViT +   DeepLabV3 Head & 5.84   & 0.861 ± 0.269 & 0.891 ± 0.240 & 0.882 ± 0.246 & 0.909 ± 0.219 \\ \hline
\multicolumn{1}{c|}{\multirow{2}{*}{L\_16\_imagenet1k}} & ViT + DeepLabV3 Head       & 310.66 & 0.862 ± 0.280 & 0.888 ± 0.258 & 0.882 ± 0.263 & 0.903 ± 0.246 \\
\multicolumn{1}{c|}{} &
  LoraViT +   DeepLabV3 Head &
  7.75 &
  \textbf{0.863 ± 0.269} &
  \textbf{0.891 ± 0.241} &
  \textbf{0.884 ± 0.243} &
  \textbf{0.911 ± 0.215} \\ \hline
\multicolumn{1}{c|}{\multirow{2}{*}{B\_32\_imagenet1k}} & ViT + DeepLabV3 Head       & 93.15 & 0.823 ± 0.325 & 0.849 ± 0.303 & 0.839 ± 0.305 & 0.866 ± 0.282 \\
\multicolumn{1}{c|}{}                                   & LoraViT +   DeepLabV3 Head & 5.84  & 0.839 ± 0.297 & 0.869 ± 0.269 & 0.855 ± 0.285 & 0.882 ± 0.260 \\ \hline
\multicolumn{1}{c|}{\multirow{2}{*}{L\_32\_imagenet1k}} & ViT + DeepLabV3 Head       & 312.87 & 0.821 ± 0.327 & 0.847 ± 0.307 & 0.849 ± 0.297 & 0.875 ± 0.276 \\
\multicolumn{1}{c|}{}                                   & LoraViT +   DeepLabV3 Head & 7.75   & 0.838 ± 0.299 & 0.868 ± 0.272 & 0.858 ± 0.279 & 0.886 ± 0.254 \\ \hline
\end{tabular}%
}
\label{Tab:tab2}
\end{table*}

\begin{table*}[h]
\centering
\caption{Crop semantic segmentation test results in terms of Jaccard Index (IoU) and F\textsubscript{1} score for the different architectures with DSTL image set}\label{tab2}
\resizebox{\textwidth}{!}{%
\begin{tabular}{cc|c|cccc}
\multicolumn{1}{l}{\textbf{}} &
  \multicolumn{1}{l|}{} &
  \multirow{2}{*}{\textbf{Trainable Parameters(M)}} &
  \multicolumn{2}{c}{\textbf{RGB}} &
  \multicolumn{2}{c}{\textbf{NIR}} \\ \cline{4-7} 
\textbf{Backbone}                                       & \textbf{Model}             &         & \textbf{IoU}  & \textbf{F1}   & \textbf{IoU}  & \textbf{F1}   \\ \hline
\multicolumn{1}{c|}{Xception}                           & DeepLabV3                  & 58.80   & 0.858 ± 0.322 & 0.869 ± 0.316 & 0.867 ± 0.302 & 0.882 ± 0.291 \\ \hline
\multicolumn{1}{c|}{\multirow{2}{*}{B\_16\_imagenet1k}} & ViT + DeepLabV3 Head       & 91.49  & 0.873 ± 0.298 & 0.891 ± 0.285 & 0.884 ± 0.290 & 0.895 ± 0.281 \\
\multicolumn{1}{c|}{}                                   & LoraViT +   DeepLabV3 Head & 5.84   & 0.876 ± 0.289 & 0.891 ± 0.278 & 0.896 ± 0.271 & 0.908 ± 0.261 \\ \hline
\multicolumn{1}{c|}{\multirow{2}{*}{L\_16\_imagenet1k}} & ViT + DeepLabV3 Head       & 310.66 & 0.878 ± 0.296 & 0.890 ± 0.287 & 0.887 ± 0.282 & 0.899 ± 0.272 \\
\multicolumn{1}{c|}{} &
  LoraViT +   DeepLabV3 Head &
  7.75 &
  \textbf{0.886 ± 0.283} &
  \textbf{0.898 ± 0.273} &
  \textbf{0.900 ± 0.266} &
  \textbf{0.912 ± 0.254} \\ \hline
\multicolumn{1}{c|}{\multirow{2}{*}{B\_32\_imagenet1k}} & ViT + DeepLabV3 Head       & 93.15 & 0.860 ± 0.308 & 0.876 ± 0.296 & 0.871 ± 0.297 & 0.886 ± 0.285 \\
\multicolumn{1}{c|}{}                                   & LoraViT +   DeepLabV3 Head & 5.84  & 0.874 ± 0.298 & 0.887 ± 0.288 & 0.888 ± 0.273 & 0.903 ± 0.257 \\ \hline
\multicolumn{1}{c|}{\multirow{2}{*}{L\_32\_imagenet1k}} & ViT + DeepLabV3 Head       & 312.87 & 0.871 ± 0.304 & 0.883 ± 0.293 & 0.876 ± 0.298 & 0.888 ± 0.288 \\
\multicolumn{1}{c|}{}                                   & LoraViT +   DeepLabV3 Head & 7.75   & 0.871 ± 0.303 & 0.883 ± 0.296 & 0.887 ± 0.276 & 0.902 ± 0.263 \\ \hline
\end{tabular}%
}
\label{Tab:tab3}
\end{table*}

Experiments employ the ViT base (B) and ViT large (L) backbones after pre-training on ImageNet-21k and fine-tuning on ImageNet-1k, where Table \ref{Tab:tab1} shows their details. In ImageNet, because the training is performed only on RGB images, the backbone networks represent only the features of the RGB domain.

Table \ref{Tab:tab2} shows semantic segmentation test results for the tree class of the RIT-18 image set using RGB and NIR images, represented in terms of the Jaccard index and $F_{1}$ score. When using Lora with the pre-trained ViT-L/16 encoder, this model outperforms all other models for the downstream task in the NIR domain.  The Jaccard index result using ViT-L/16 with Lora for the NIR domain demonstrates that even though the number of parameters decreases by approximately 97.5\%, it performs 0.2\% better than the ViT-L/16 without Lora. Using Lora with the ViT-L/16 backbone shows an increase in performance by 2.1\% for the NIR domain compared to the RGB domain. 

\begin{figure}[h]
    \centering
    
    \subfloat[]{
        \includegraphics*[width=0.17\textwidth]{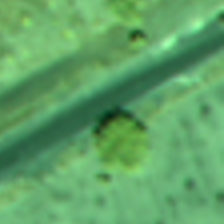}
    } \negthinspace
    \subfloat[]{
        \includegraphics*[width=0.17\textwidth]{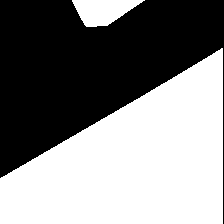}
    } \negthinspace    
    \subfloat[]{
        \includegraphics*[width=0.17\textwidth]{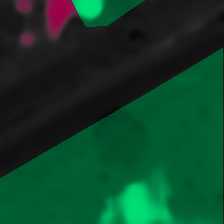}
    } \negthinspace
    \subfloat[]{
        \includegraphics*[width=0.17\textwidth]{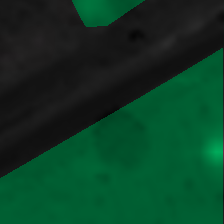}
    } \negthinspace
    \subfloat[]{
        \includegraphics*[width=0.17\textwidth]{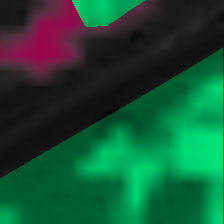}
    } 
    
    \vspace{0.1cm}
    
    \subfloat{
        \includegraphics*[width=0.17\textwidth]{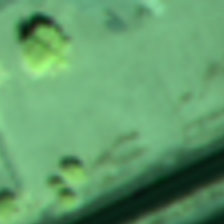}
    } \negthinspace
    \subfloat{
        \includegraphics*[width=0.17\textwidth]{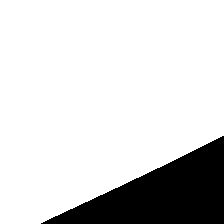}
    } \negthinspace    
    \subfloat{
        \includegraphics*[width=0.17\textwidth]{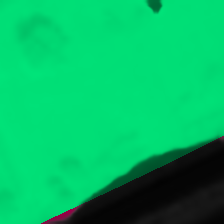}
    } \negthinspace
    \subfloat{
        \includegraphics*[width=0.17\textwidth]{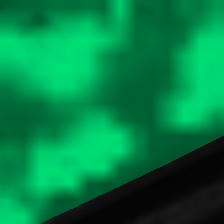}
    } \negthinspace
    \subfloat{
        \includegraphics*[width=0.17\textwidth]{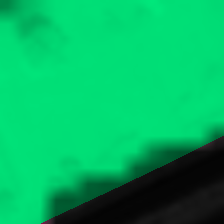}
    } 
    
    \vspace{0.1cm}
    
    \subfloat{
        \includegraphics*[width=0.17\textwidth]{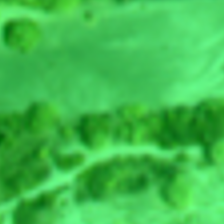}
    } \negthinspace
    \subfloat{
        \includegraphics*[width=0.17\textwidth]{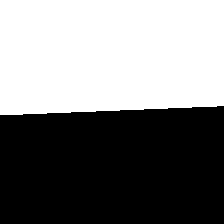}
    } \negthinspace    
    \subfloat{
        \includegraphics*[width=0.17\textwidth]{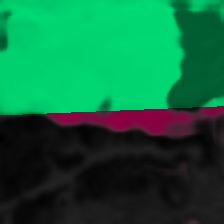}
    } \negthinspace
    \subfloat{
        \includegraphics*[width=0.17\textwidth]{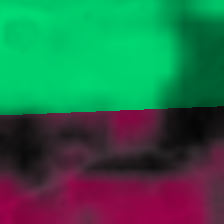}
    } \negthinspace
    \subfloat{
        \includegraphics*[width=0.17\textwidth]{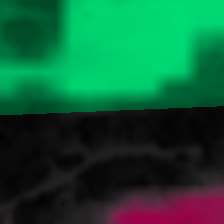}
    } 
    
    \vspace{0.1cm}
    
    \subfloat{
        \includegraphics*[width=0.17\textwidth]{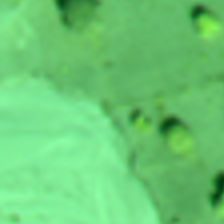}
    } \negthinspace
    \subfloat{
        \includegraphics*[width=0.17\textwidth]{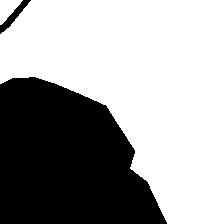}
    } \negthinspace    
    \subfloat{
        \includegraphics*[width=0.17\textwidth]{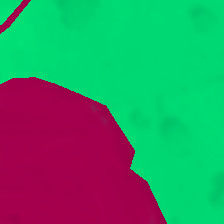}
    } \negthinspace
    \subfloat{
        \includegraphics*[width=0.17\textwidth]{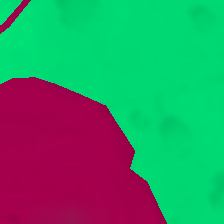}
    } \negthinspace
    \subfloat{
        \includegraphics*[width=0.17\textwidth]{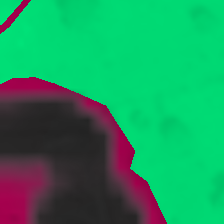}
    }   
    \caption{Crop predictions for DSTL image set in NIR domain. Light green represents a hit, dark green represents a miss, and red represents a false alarm.(a) Image. (b) Ground-truth mask. (c) DeepLabV3. (d) ViT (L$-$16). (e) LoraViT (L$-$16).}
    \label{fig:predictions1}
\end{figure}

\begin{figure}[h]
    \centering
    
    \subfloat[]{
        \includegraphics*[width=0.17\textwidth]{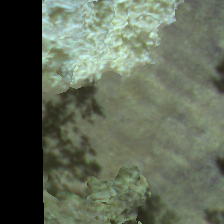}
    } \negthinspace
    \subfloat[]{
        \includegraphics*[width=0.17\textwidth]{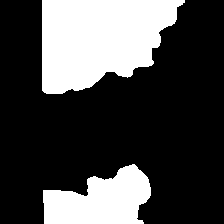}
    } \negthinspace    
    \subfloat[]{
        \includegraphics*[width=0.17\textwidth]{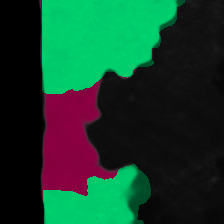}
    } \negthinspace
    \subfloat[]{
        \includegraphics*[width=0.17\textwidth]{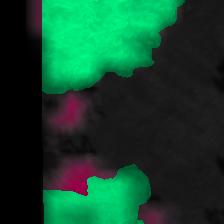}
    } \negthinspace
    \subfloat[]{
        \includegraphics*[width=0.17\textwidth]{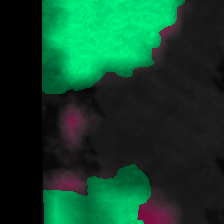}
    } 
    
    \vspace{0.1cm}
    
    \subfloat{
        \includegraphics*[width=0.17\textwidth]{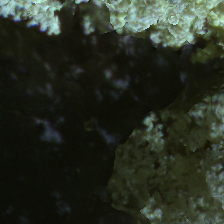}
    } \negthinspace
    \subfloat{
        \includegraphics*[width=0.17\textwidth]{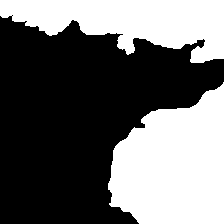}
    } \negthinspace    
    \subfloat{
        \includegraphics*[width=0.17\textwidth]{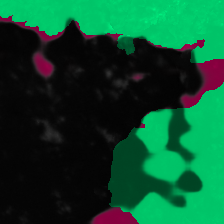}
    } \negthinspace
    \subfloat{
        \includegraphics*[width=0.17\textwidth]{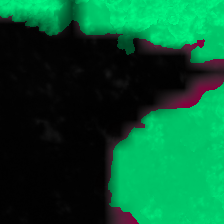}
    } \negthinspace
    \subfloat{
        \includegraphics*[width=0.17\textwidth]{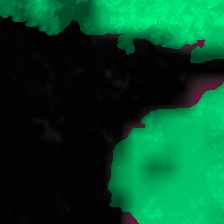}
    } 
    
    \vspace{0.1cm}
    
    \subfloat{
        \includegraphics*[width=0.17\textwidth]{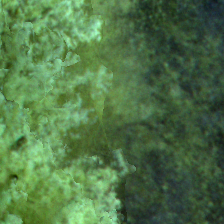}
    } \negthinspace
    \subfloat{
        \includegraphics*[width=0.17\textwidth]{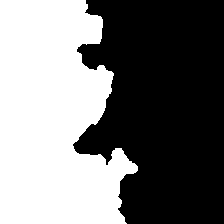}
    } \negthinspace    
    \subfloat{
        \includegraphics*[width=0.17\textwidth]{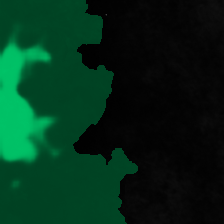}
    } \negthinspace
    \subfloat{
        \includegraphics*[width=0.17\textwidth]{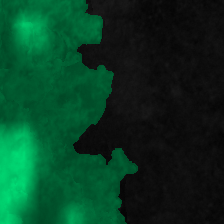}
    } \negthinspace
    \subfloat{
        \includegraphics*[width=0.17\textwidth]{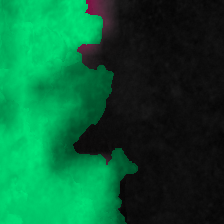}
    } 
    
    \vspace{0.1cm}
    
    \subfloat{
        \includegraphics*[width=0.17\textwidth]{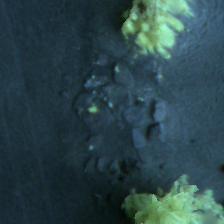}
    } \negthinspace
    \subfloat{
        \includegraphics*[width=0.17\textwidth]{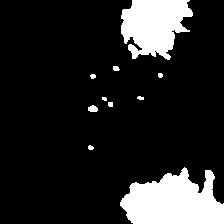}
    } \negthinspace    
    \subfloat{
        \includegraphics*[width=0.17\textwidth]{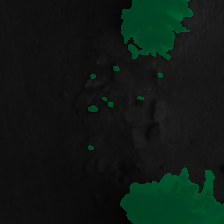}
    } \negthinspace
    \subfloat{
        \includegraphics*[width=0.17\textwidth]{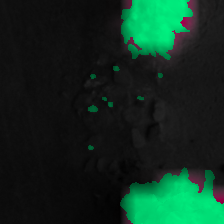}
    } \negthinspace
    \subfloat{
        \includegraphics*[width=0.17\textwidth]{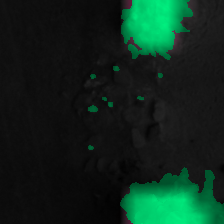}
    }   
    \caption{Tree predictions for RIT-18 image set in NIR domain. Light green represents a hit, dark green represents a miss, and red represents a false alarm. (a) Image. (b) Ground-truth mask. (c) DeepLabV3. (d) ViT (L$-$16). (e) LoraViT (L$-$16).}
    \label{fig:predictions2}
\end{figure}

An attractive feature of the low-rank adaptation is that despite a significant distributional difference between the RGB and NIR domains, the performances of backbones with Lora in the NIR domain are superior to RGB counterparts across all results. Besides that, the Lora decreases the total number of trainable parameters by approximately 97\%. 

Additionally, the Lora-based ViT-L/16 backbone competes with the state-of-the-art DeepLabV3 and outperforms the Jaccard index performance of this model by approximately 2.7\% in the NIR domain. As expected, larger ViT models like ViT-L/16 demonstrate significantly stronger intermediate representations when large pretraining datasets like ImageNet-21k are used \cite{raghu2021vision}.

As seen in Table \ref{Tab:tab3}, adapting the ViT-L/16 backbone with Lora in the NIR domain shows the best performance also on the DSTL image set for crop objects. The experimental results show that using ViT-L/16 with Lora on NIR images improves the Jaccard index performance by 1.3\% compared to employing ViT-L/16 without Lora. Adapting the ViT-L/16 backbone to the NIR domain with Lora shows a 1.4\% improvement in Jaccard index performance compared to its adaptation to the RGB domain. As shown in Table \ref{Tab:tab2} and Table \ref{Tab:tab3},  a patch size of $16$ generally performs better than those with a patch size of $32$, which conforms with the findings in the literature that a smaller patch size performs better \cite{caron2021emerging, dosovitskiy2020image}.

Fig. \ref{fig:predictions1} and Fig. \ref{fig:predictions2} present visual comparisons of the various ViT-L/16 backbone cases for DSTL and RIT18 image sets, respectively. ViT-L/16 backbone with Lora in the NIR domain shows a lower false alarm rate than other cases. Additionally, there are fewer missed detections when capturing small target objects and complex details.

\section{Conclusion}
This study explores the potential benefits of using pre-trained ViT backbones in the RGB domain with low-rank adaptation for downstream tasks in the NIR domain. Due to the domain shift problem, directly using ViT backbones or traditional semantic segmentation models does not yield sufficient results in the NIR domain. However, extensive experiments in this study demonstrate that using LoRA with pre-trained ViT backbones provides the best performance for downstream tasks applied to NIR images. Lora is advantageous, as it typically contributes to the overall stability of the domain adaptation process.

Given the scarcity of large amounts of labeled data for NIR bands that contain discriminative information for vegetation, this study is a promising approach to addressing the problem of insufficient utilization of NIR bands in the semantic segmentation process. For future studies, it may be valuable to investigate the application of LoRA to segmentation architectures for the NIR domain using different segmentation heads.




%


\bibliographystyle{plain}

\bibliography{Mybib}

\begin{thebibliography}{10}

\bibitem{arogoundade2023leveraging}
Adeola~M Arogoundade, Onisimo Mutanga, John Odindi, and Omosalewa Odebiri.
\newblock Leveraging google earth engine to estimate foliar c: N ratio in an african savannah rangeland using sentinel 2 data.
\newblock {\em Remote Sensing Applications: Society and Environment}, 30:100981, 2023.

\bibitem{aslahishahri2021rgb}
Masoomeh Aslahishahri, Kevin~G Stanley, Hema Duddu, Steve Shirtliffe, Sally Vail, Kirstin Bett, Curtis Pozniak, and Ian Stavness.
\newblock From rgb to nir: Predicting of near infrared reflectance from visible spectrum aerial images of crops.
\newblock In {\em Proceedings of the IEEE/CVF International Conference on Computer Vision}, pages 1312--1322, 2021.

\bibitem{caron2021emerging}
Mathilde Caron, Hugo Touvron, Ishan Misra, Herv{\'e} J{\'e}gou, Julien Mairal, Piotr Bojanowski, and Armand Joulin.
\newblock Emerging properties in self-supervised vision transformers.
\newblock In {\em Proceedings of the IEEE/CVF international conference on computer vision}, pages 9650--9660, 2021.

\bibitem{chen2017deeplab}
Liang-Chieh Chen, George Papandreou, Iasonas Kokkinos, Kevin Murphy, and Alan~L Yuille.
\newblock Deeplab: Semantic image segmentation with deep convolutional nets, atrous convolution, and fully connected crfs.
\newblock {\em IEEE transactions on pattern analysis and machine intelligence}, 40(4):834--848, 2017.

\bibitem{dong2024upetu}
Zhe Dong, Yanfeng Gu, and Tianzhu Liu.
\newblock Upetu: A unified parameter-efficient fine-tuning framework for remote sensing foundation model.
\newblock {\em IEEE Transactions on Geoscience and Remote Sensing}, 2024.

\bibitem{dosovitskiy2020image}
Alexey Dosovitskiy, Lucas Beyer, Alexander Kolesnikov, Dirk Weissenborn, Xiaohua Zhai, Thomas Unterthiner, Mostafa Dehghani, Matthias Minderer, Georg Heigold, Sylvain Gelly, et~al.
\newblock An image is worth 16x16 words: Transformers for image recognition at scale.
\newblock {\em arXiv preprint arXiv:2010.11929}, 2020.

\bibitem{gkillas2023cost}
Alexandros Gkillas, Dimitrios Kosmopoulos, and Kostas Berberidis.
\newblock Cost-efficient coupled learning methods for recovering near-infrared information from rgb signals: Application in precision agriculture.
\newblock {\em Computers and Electronics in Agriculture}, 209:107833, 2023.

\bibitem{he2023sensitivity}
Haoyu He, Jianfei Cai, Jing Zhang, Dacheng Tao, and Bohan Zhuang.
\newblock Sensitivity-aware visual parameter-efficient fine-tuning.
\newblock In {\em Proceedings of the IEEE/CVF International Conference on Computer Vision}, pages 11825--11835, 2023.

\bibitem{hu2021lora}
Edward~J Hu, Yelong Shen, Phillip Wallis, Zeyuan Allen-Zhu, Yuanzhi Li, Shean Wang, Lu~Wang, and Weizhu Chen.
\newblock Lora: Low-rank adaptation of large language models.
\newblock {\em arXiv preprint arXiv:2106.09685}, 2021.

\bibitem{hu2024airs}
Leiyi Hu, Hongfeng Yu, Wanxuan Lu, Dongshuo Yin, Xian Sun, and Kun Fu.
\newblock Airs: Adapter in remote sensing for parameter-efficient transfer learning.
\newblock {\em IEEE Transactions on Geoscience and Remote Sensing}, 2024.

\bibitem{li2023enhancing}
Yaqin Li, Dandan Wang, Cao Yuan, Hao Li, and Jing Hu.
\newblock Enhancing agricultural image segmentation with an agricultural segment anything model adapter.
\newblock {\em Sensors}, 23(18):7884, 2023.

\bibitem{liu2022polyhistor}
Yen-Cheng Liu, Chih-Yao Ma, Junjiao Tian, Zijian He, and Zsolt Kira.
\newblock Polyhistor: Parameter-efficient multi-task adaptation for dense vision tasks.
\newblock {\em Advances in Neural Information Processing Systems}, 35:36889--36901, 2022.

\bibitem{liu2021swin}
Ze~Liu, Yutong Lin, Yue Cao, Han Hu, Yixuan Wei, Zheng Zhang, Stephen Lin, and Baining Guo.
\newblock Swin transformer: Hierarchical vision transformer using shifted windows.
\newblock In {\em Proceedings of the IEEE/CVF international conference on computer vision}, pages 10012--10022, 2021.

\bibitem{ma2024transfer}
Yuchi Ma, Shuo Chen, Stefano Ermon, and David~B Lobell.
\newblock Transfer learning in environmental remote sensing.
\newblock {\em Remote Sensing of Environment}, 301:113924, 2024.

\bibitem{olimov2021ref}
Bekhzod Olimov, Jeonghong Kim, and Anand Paul.
\newblock Ref-net: robust, efficient, and fast network for semantic segmentation applications using devices with limited computational resources.
\newblock {\em IEEE Access}, 9:15084--15098, 2021.

\bibitem{raghu2021vision}
Maithra Raghu, Thomas Unterthiner, Simon Kornblith, Chiyuan Zhang, and Alexey Dosovitskiy.
\newblock Do vision transformers see like convolutional neural networks?
\newblock {\em Advances in neural information processing systems}, 34:12116--12128, 2021.

\bibitem{ulku2022deep}
Irem Ulku, Erdem Akag{\"u}nd{\"u}z, and Pedram Ghamisi.
\newblock Deep semantic segmentation of trees using multispectral images.
\newblock {\em IEEE Journal of Selected Topics in Applied Earth Observations and Remote Sensing}, 15:7589--7604, 2022.

\bibitem{xue2024adapting}
Bowei Xue, Han Cheng, Qingqing Yang, Yi~Wang, and Xiaoning He.
\newblock Adapting segment anything model to aerial land cover classification with low rank adaptation.
\newblock {\em IEEE Geoscience and Remote Sensing Letters}, 2024.

\bibitem{zhang2023two}
Ting Zhang, Haijian Shen, Sadaqat ur~Rehman, Zhaoying Liu, Yujian Li, and Obaid ur~Rehman.
\newblock Two-stage domain adaptation for infrared ship target segmentation.
\newblock {\em IEEE Transactions on Geoscience and Remote Sensing}, 2023.

\end{thebibliography}

\end{document}